\documentclass[runningheads]{llncs}
\usepackage{graphicx}
\usepackage{amsmath}
\usepackage{amsfonts}
\usepackage{booktabs}
\usepackage{multirow}
\usepackage{hyperref}

\usepackage{paralist}
\usepackage{xcolor}

\begin{document}
\title{On robustness of generative representations against catastrophic forgetting}
\author{Wojciech Masarczyk\inst{1}\orcidID{0000-0001-6548-4139}\\
Kamil Deja\inst{1}\thanks{Work done prior joining Amazon}\orcidID{0000-0003-1156-5544}\\
Tomasz Trzcinski\inst{1,2,3}\orcidID{0000-0002-1486-8906}}
\authorrunning{W. Masarczyk et al.}
\institute{Warsaw University of Technology \and Jagiellonian University \and Tooploox \\
%   \email{\{firstname.lastname\}@pw.edu.pl}}
  \email\{wojciech.masarczyk.dokt, kamil.deja.dokt, tomasz.trzcinski\}@pw.edu.pl}
\maketitle              % typeset the header of the contribution
\begin{abstract}

Catastrophic forgetting of previously learned knowledge while learning new tasks is a widely observed limitation of contemporary neural networks. Although many continual learning methods are proposed to mitigate this drawback, the main question remains unanswered: what is the root cause of catastrophic forgetting? In this work, we aim at answering this question by posing and validating a set of research hypotheses related to the specificity of representations built internally by neural models. More specifically, we design a set of empirical evaluations that compare the robustness of representations in discriminative and generative models against catastrophic forgetting. We observe that representations learned by discriminative models are more prone to catastrophic forgetting than their generative counterparts, which sheds new light on the advantages of developing generative models for continual learning. Finally, our work opens new research pathways and possibilities to adopt generative models in continual learning beyond mere replay mechanisms.

\end{abstract}
\section{Introduction}
Neural networks are widely used across many real-life applications, ranging from image recognition~\cite{ILSVRC15} to natural language processing~\cite{vaswani2017attention}. Nevertheless, neural models used in those applications assume identical and independently distributed training data - the assumption rarely met in practice. As a result, contemporary neural network models are prone to catastrophic forgetting~\cite{1999french} - a well-known limitation of neural networks that results in the erosion of previously learned knowledge. 
Continual learning is a field of machine learning that aims at addressing this pitfall of neural models by constant adaption to new data. 
The majority of works in this field focus on developing methods for mitigating the effects of catastrophic forgetting~\cite{parisi2018review}. These methods can be grouped into three categories -- regularization based~\cite{2017zenke+2,2017kirkpatrick+many}, rehearsal methods~\cite{van2018generative,caccia2020online,von2019continual,2020deja+4,2017shin+3,2019rolnick+4} and methods using dynamic architectures~\cite{2018masse+2,2018mallya+1,2019golkar+2,2018mallya+2,2020wortsman+6,2016rusu+7,2017yoon+3}. Nevertheless, recent findings~\cite{prabhu2020greedy,thai2021continual} show that it is possible to surpass well-established continual learning methods across popular benchmarks with heuristic-based baselines. The surprising effectiveness of these baselines indicates that the root cause of catastrophic forgetting is yet to be discovered, and we follow this intuition in our work.

\begin{figure}[t!]
     \centering
     \includegraphics[width=0.95\linewidth]{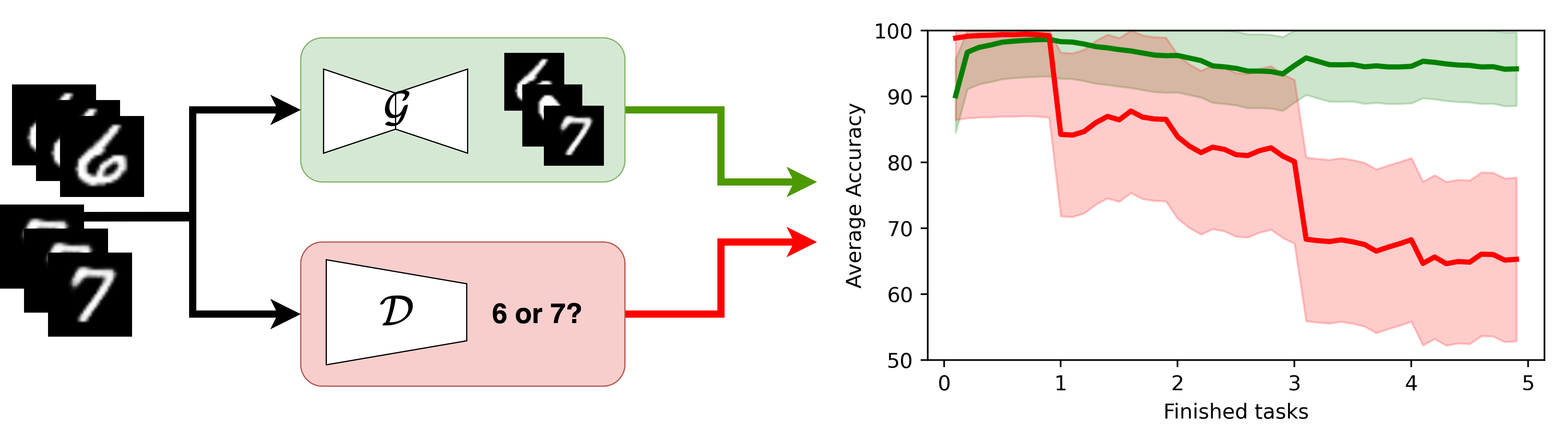}
     \caption{A schematic overview of our main experiment. Representations learned by the generative model are less susceptible to catastrophic forgetting than discriminative ones. Results of the main experiment and details can be found at Sec.~\ref{sec:classification}.}.
          \vspace{-0.4cm}\label{Fig:teaser}
     \vspace{-0.4cm}
\end{figure}

We investigate the catastrophic forgetting with methodological rigor; we state and empirically validate research hypotheses that shed new light on this phenomenon. We build upon the works of~\cite{davidson2020sequential,ramasesh2021anatomy} where the roots of forgetting are analyzed and the findings of~\cite{thai2021continual,verwimp2021rehearsal} that analyze the effectiveness of intuitive solutions to this problem. Here, contrary to previous works, we postulate to look at the continual learning from the perspective of internal representations of neural networks and analyze its impact on the final performance of the continually learned model. 
Inspired by~\cite{thai2021continual}, we argue that the dynamics of catastrophic forgetting depends on the task at hand, yet contrary to this work, we analyze this observation from the perspective of internal neural representations rather than peculiarities of continual learning tasks.

To summarize, the main contribution of our work is the statement of the following research hypotheses, along with their empirical validation:\\
\noindent
\textbf{Hypothesis 1} -- \textit{Representations learned by autoencoders and variational autoencoders are less prone to catastrophic forgetting than representations of discriminative models.}

\noindent
\textbf{Hypothesis 2} -- \textit{Autoencoders and variational autoencoders learn more transferable features than discriminative models.}

Moreover, the results from our experiments provide an explanation for the effect observed in~\cite{thai2021continual}, where the authors argue that continual reconstruction tasks do not suffer from catastrophic forgetting.
Our experiments show that this observation can be explained from the perspective of generative representations~\footnote{Throughout this work, we refer to the hidden representations of autoencoders or variational autoencoders as generative representations. Similarly, we refer to the reconstruction task as the generative task. Although autoencoders are not strictly generative models, we use this term to highlight the difference in the objective of the particular model. In the case of the reconstruction task, the aim is to \textit{generate} a sample, while in the case of classification, the aim is to \textit{discriminate} samples.}. This suggests that the lack of catastrophic forgetting is not exclusively linked to the continual reconstruction task.

Last but not least, our experiments show that it is possible to achieve the performance of over $90\%$ average accuracy on popular continual learning benchmarks without any specific mechanism to overcome catastrophic forgetting. This raises the question of whether these datasets and currently used evaluation protocols should be used for benchmarking novel continual learning methods as they no longer pose any significant challenge, as the results in Table~\ref{Tab:after_first_task} show.

\noindent

\section{Related works}
Following~\cite{parisi2018review}, continual learning methods can be grouped into three categories: regularization, dynamic architectures, and rehearsal.   

\paragraph{\textbf{Regularization}}

These methods typically train a~single model on the subsequent tasks and impose specifically defined regularization techniques which penalize the change of important parameters of the model~\cite{2017zenke+2,2017kirkpatrick+many}.

\paragraph{\textbf{Dynamic architectures}}
Here, the solution comprises many substructures, which are usually trained in isolation for specific tasks. On top of these structures, a separate mechanism decides which substructure to use during evaluation. In~\cite{2016rusu+7,2017yoon+3,2018xu+1} new structural elements are added to the model for each new task which requires growing memory for the whole model. 
In \cite{2018masse+2,2018mallya+1,2019golkar+2,2018mallya+2,2020wortsman+6} a~large model is considered from which independent submodels are selected for subsequent tasks.

\paragraph{\textbf{Rehearsal}}
Methods in this group try to prevent forgetting by retraining the model with a combination of new and previous data examples. 
Methods presented in \cite{2019rolnick+4,aljundi2019online} employ a~memory buffer to store all or possibly most relevant previous data examples. To overcome the scalability issues of the standard buffer approach, authors~\cite{2017shin+3} propose to replace the buffer with a~generative model. This method introduced a general schema that was further extended in several new approaches with various generative models~\cite{van2018generative,caccia2020online,von2019continual,2020deja+4}.

\paragraph{}
While most of the works develop new methods that are more robust to catastrophic forgetting, only several works investigate the actual phenomenon~\cite{davidson2020sequential,nguyen2020dissecting,nguyen2019understanding,ramasesh2021anatomy}.
Specifically, in~\cite{ramasesh2021anatomy} authors analyze the relationship between semantic similarity of tasks and magnitude of catastrophic forgetting. %on the model. 
In~\cite{davidson2020sequential} authors show that the effect of catastrophic forgetting diminishes with the prolonged exposure to the domain, which suggests that catastrophic forgetting could be an artifact of immature systems. Another approach~\cite{nguyen2020dissecting} investigates the problem with the tools of Explainable Artificial Intelligence (XAI), comparing the effects of catastrophic forgetting for different layers of CNN.

%Recently, in~\cite{gallardo2021selfsupervised} authors analyzed the impact of self-supervised pretraining for continual learning. Authors argue that representations learned within self-supervised training are task agnostic and hence superior to the counterparts obtained within the supervised pretraining. The effect is even more substantial in small dataset regimes.
Our work is directly influenced by Thai et al.~\cite{thai2021continual}, where authors claim that networks trained in continual reconstruction tasks do not suffer from catastrophic forgetting.
We show that different dynamics of forgetting in reconstruction and classification tasks are linked to the representations of the data constructed by the neural networks. Specifically, we argue that learning representations through generative modeling is naturally more aligned with continual learning. 

\section{Methodology}

\begin{figure}[!t]
     \centering
     \includegraphics[width=\linewidth]{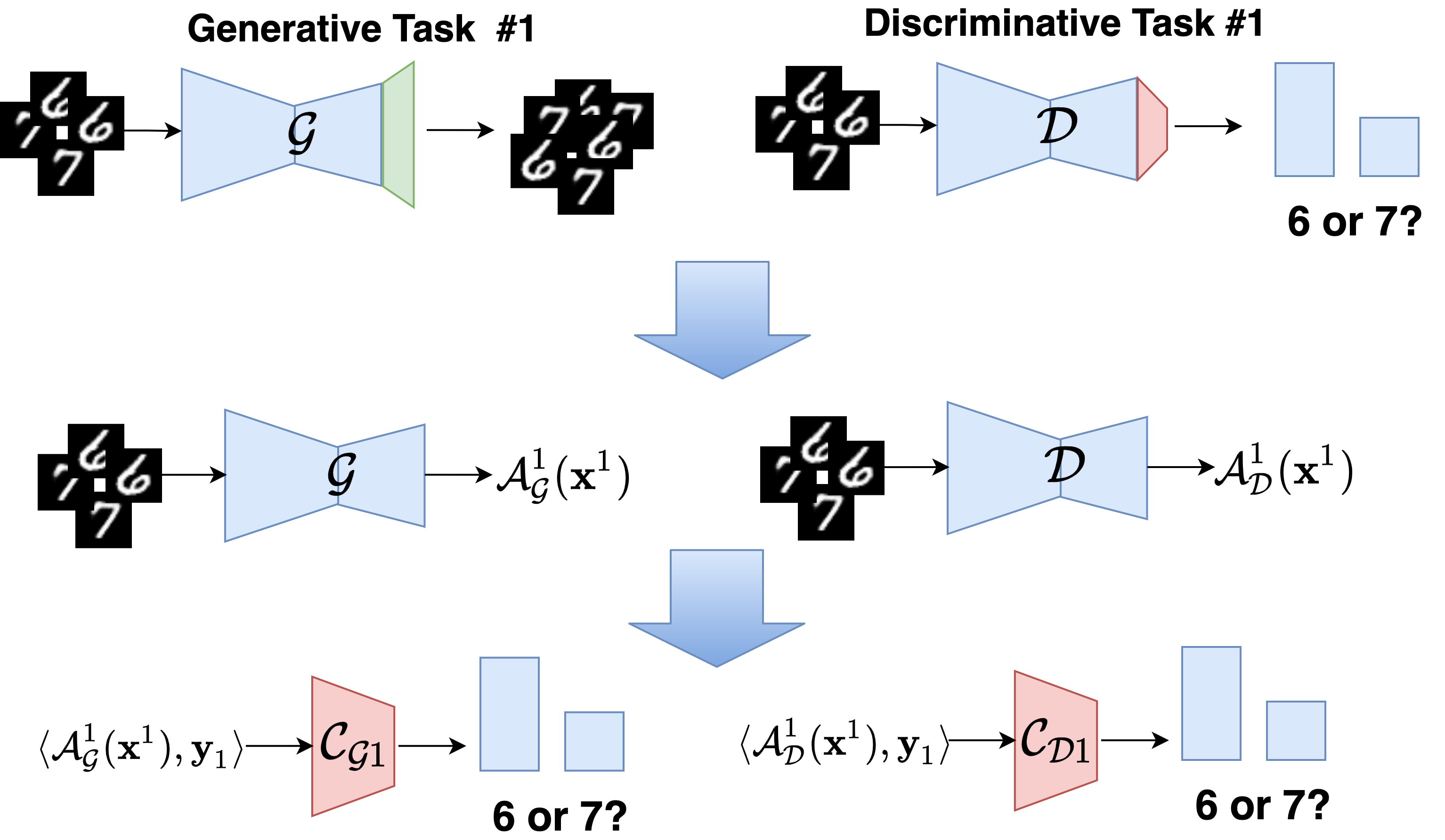}
     \caption{Schematic overview describing the process of learning and collecting discriminative and generative representations.}\label{Fig:exp1_explanation}
\end{figure}

To examine and compare representations from different models, we need to design a fair method to learn these models and collect respective representations. To that end we consider three types of models: discriminative -- $\mathcal{D}$, generative based on Autoencoder -- $\mathcal{G}_{AE}$ and generative based on Variational Autonecoder~\cite{kingma2014vae} -- $\mathcal{G}_{VAE}$. To make a fair comparison, these networks share the same architecture, except for the last layer, defined by the model's objective. In the case of the discriminative task, the last layer has output neurons to discriminate between classes. In the generative task, the final layer outputs vectors of equal size as input data. Depending on the model, we train the networks with different objective functions. $\mathcal{D}$ model is trained with CrossEntropy, $\mathcal{G}_{AE}$ minimizes the MSE and the $\mathcal{G}_{VAE}$ uses ELBO. As shown in Fig.~,\ref{Fig:exp1_explanation} we train all models on the same training sequence $T_{N}$ but with different objectives. The index of particular task is denoted by $k$, where $k = 1, \ldots, N$. To obtain the representations of data for a particular model, we feed the data to the model and collect the representations from the penultimate layer of the model.  We use $\mathcal{A}^{t}_{\mathcal{D}}(\mathbf{x}^{k})$, to denote activations from model $\mathcal{D}$ after finishing task $t$ for input data $ \langle \mathbf{x}^{k} \rangle$ from task $k$.

\paragraph{\textbf{Datasets}}
We use three commonly used continual learning benchmarks to create a continuous sequence of tasks: splitMNIST, splitFashionMNIST, splitCIFAR. We follow the typical formulation of the continual learning task and split the datasets to $5$ disjoint subsets with two classes for each task. 

\paragraph{\textbf{Architectures}}
We use simple autoencoder architecture composed of three encoding fully connected layers and three decoding fully connected layers for all the tasks, followed by ReLU activations except for the last layer. For the generative task, the output layer is a single fully connected layer that maps the penultimate layer's activations to the vector of equal size with input data. For the classification task, we use a multi-head output layer with two neurons per task. 
For MNIST and FashionMNIST datasets, we use architecture with (512, 256, 8, 256, 512) neurons on each hidden layer. For the CIFAR dataset, we use the same architecture with a bottleneck of 128 neurons.

\subsection{Hypothesis 1}

\subsubsection{Discriminative abilities of representations}~\label{sec:classification}

First, to test our hypothesis, we look at the problem of catastrophic forgetting from the accuracy perspective, as it is the most widely adopted measure of the catastrophic forgetting phenomenon.
To measure the robustness of representations to the catastrophic forgetting, we train the models $\mathcal{D}$, $\mathcal{G}_{AE}$ and $\mathcal{G}_{VAE}$ on identical data sequences $T_{N}$ and collect their representations from penultimate layer for all data splits. These representations are used to train a set of linear classifiers. Although we collect the representations for all data splits throughout the training, we train respective classifiers only when the corresponding task is active. This means that we train only the first classifier during the first task, and the rest remain untrained. After finishing the first task, we freeze the first classifier for the rest of the training. This way, degradation of its performance can be attributed exclusively to forgetting the representations for the first task, and we can directly measure the amount of forgetting in the model.

More precisely, after each epoch of training, for task data of the form $\langle \mathbf{x}^{k}, \mathbf{y}^{k} \rangle$, where $k$ denotes the respective data split, we collect the representations for model $\mathcal{D}$ from its penultimate layer $\mathcal{A}^{k}_{\mathcal{D}}(\mathbf{x}^{j})$ for $j = k$ and train softmax regression classifier $\mathcal{C}_{\mathcal{D}j}$ on dataset of the form $\langle \mathcal{A}^{k}_{\mathcal{D}}(\mathbf{x}^{k}), \mathbf{y}^{k} \rangle$, where $k$ denotes the present task. 
Next, the trained softmax classifiers are evaluated on the validation dataset after extracting features with respective backbones. 
Note that in the above approach, classifiers $\mathcal{C}_{\mathcal{D}j}$ for $j > k$, remain untrained. 
Since the classifiers $\mathcal{C}_{\mathcal{D}j}$ are frozen after completing $j$-th task, the loss of performance may only be attributed to the drift of representations from the penultimate layer. Therefore, the smaller the loss of the classifier's performance, the more resilient the features are to the catastrophic forgetting. The same procedure is applied to the models $\mathcal{G}_{AE}$ and $\mathcal{G}_{VAE}$ as shown in Fig~\ref{Fig:exp1_explanation}.

\begin{figure}[!htb]
     \centering
     \includegraphics[width=\linewidth]{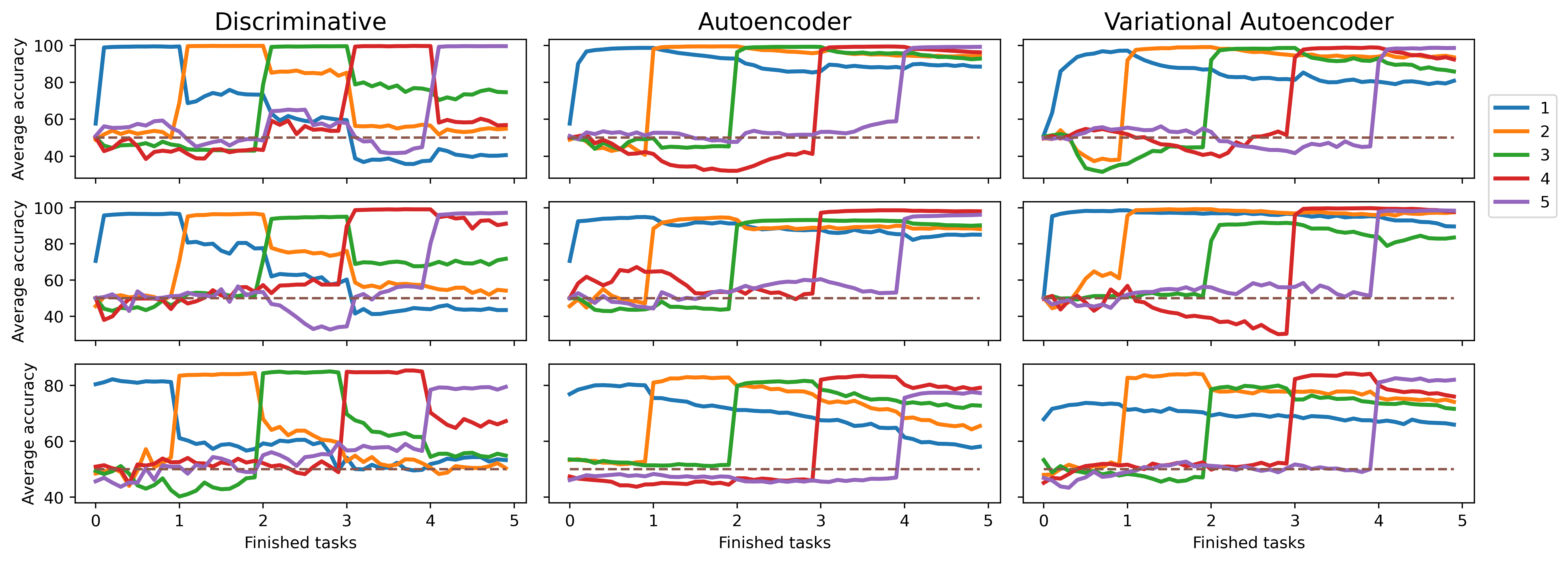}
     \caption{Results of the first experiment. Each curve represents the average accuracy on specific task with respect to the finished tasks in the continual training. The dashed line is for the reference presenting the performance of a random classifier. Columns represent results for representations of $\mathcal{D}$, $\mathcal{G}_{AE}$ and $\mathcal{G}_{VAE}$ respectively. Top row -- MNIST, middle - FashionMNIST, bottom -- CIFAR-10.}\label{Fig:main_experiment}
\end{figure}

Fig.~\ref{Fig:main_experiment} depicts the results of this experiment. Columns represent results for $\mathcal{D}$, $\mathcal{G}_{AE}$, $\mathcal{G}_{VAE}$ respectively. Starting from the top, the rows present the results for MNIST, FashionMNIST, and CIFAR. Different colors present the accuracy for different tasks.
As one can see in the left column of Fig.~\ref{Fig:main_experiment}, for all three benchmarks, the classifier's performance trained with representations of discriminative model suddenly drops after introducing new tasks and degrades further to the region of random guessing (depicted as dashed line). In contrast, for the generative case for both models, the performance is stable throughout the whole sequence of training, suggesting that the representations of tested generative models are almost immune to the problem of catastrophic forgetting, especially in the case of simpler datasets as MNIST and FashionMNIST. In the case of CIFAR-10, the amount of forgetting is considerable for the case of $\mathcal{G}_{AE}$. However, the degradation of the performance is more stable than in the case of the discriminative model. This suggests that forgetting in generative models has different nature than forgetting in discriminative models. The right column shows that the representations of $\mathcal{G}_{VAE}$ model suffer the least in the case of CIFAR-10. Explaining this difference requires further analysis, and we foresee it as future work.

Surprisingly, in MNIST and FashionMNIST, achieving an average accuracy above 90\% without any dedicated mechanism to overcome catastrophic forgetting is possible. This raises the question of whether these datasets and evaluation protocols should be used to benchmark novel methods in continual learning.

The above experiment proves the validity of the first hypothesis through the lenses of accuracy as it is a widely adopted measure to estimate catastrophic forgetting. However, such an approach may not be fully informative as generative representations can have poor discriminative abilities. To address this limitation, we propose to measure catastrophic forgetting through the index of representations similarity.

\subsubsection{Centered Kernel Alignment (CKA)}
To further examine the validity of our hypothesis that representations learned by generative models are less susceptible to catastrophic forgetting, we investigate the evolution of network representations in time. 
Since the above experiment is directly linked to the discriminative abilities of the representations, here we use a task-agnostic measure to directly estimate the drift of the representations in continual learning training.
For that purpose, we use the well-established method of Centered Kernel Alignment (CKA)~\cite{kornblith2019cka}, defined as: 
\begin{equation}
\operatorname{CKA}(X, Y)=\frac{\left\|X^{T} Y\right\|_{F}^{2}}{\left\|X^{T} X\right\|_{F}\left\|Y^{T} Y\right\|_{F}},
\end{equation}
where $X \in \mathbb{R}^{n \times m_x}$ and $Y \in \mathbb{R}^{n \times m_y}$. CKA takes values from 0 to 1, where 1 means identical representations. Using CKA, it is possible to estimate the similarity of representations obtained for different datasets or different dimensionality. We measure the similarity between representations of the same network collected at different moments of the training of the neural network.
Specifically, to analyze the evolution of representations from model $\mathcal{D}$, we collect the reference representations $\mathcal{A}^{k}_{\mathcal{D}}(\mathbf{x}^{k})$ just after finishing task $k$. Then, to measure the relative drift of representations on task $k$, we compute the CKA index between the reference and current representations $\operatorname{CKA}(\mathcal{A}^{k}_{\mathcal{D}}(\mathbf{x}^{k}), \mathcal{A}^{j}_{\mathcal{D}}(\mathbf{x}^{k}))$, for $j \geq k$. We follow this procedure for each task in training sequence $T_{N}$. The analogous procedure is applied to the representations of models $\mathcal{G}_{AE}$ and $\mathcal{G}_{VAE}$.

\begin{figure}[!htb]
     \centering
     \includegraphics[width=\linewidth]{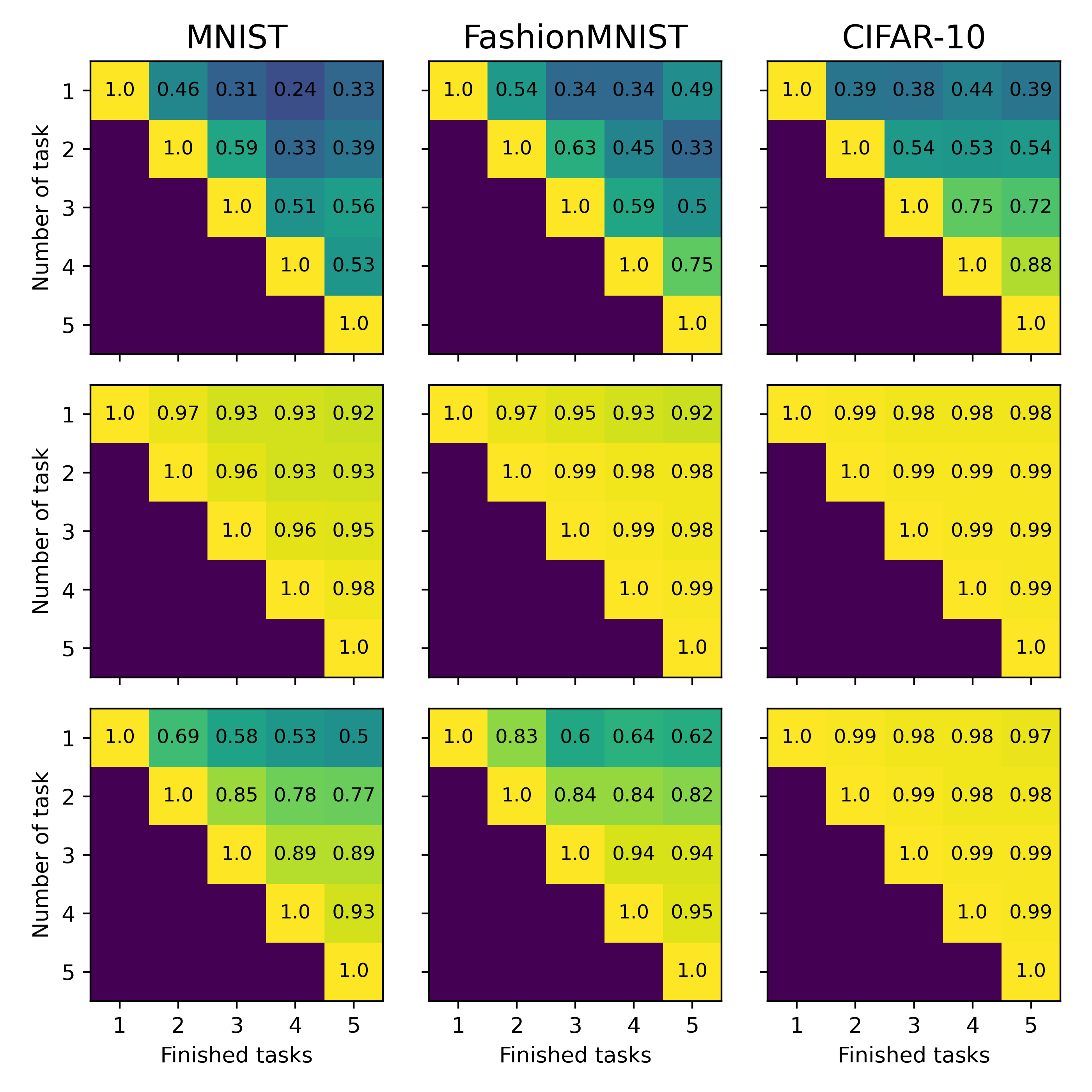}
      \caption{Illustration of the evolution of representations similarity measure with CKA index. Starting from top the rows represent results of different models: $\mathcal{D}$, $\mathcal{G}_{AE}$ and $\mathcal{G}_{VAE}$ respectively. Columns represent results on different datasets (from left to right) -- MNIST, FashionMNIST, CIFAR-10.}\label{Fig:cka}
\end{figure}

Fig.~\ref{Fig:cka} presents the results of the experiment. In the case of discriminative representations (top row), we can see an abrupt change of representations just after introducing the new task on all tested datasets. This is in line with the results from the previous experiment, which show that most of the performance is lost during the next task. In the following tasks, the representations stabilize, but these representations are no longer valuable since classifying with them is not better than random guessing (see Fig.~\ref{Fig:main_experiment}).
The results from the middle row of Fig.~\ref{Fig:cka} show that representations from $\mathcal{G}_{AE}$ slightly evolve during the training on consecutive tasks and remain similar to the reference representations. These results directly support hypothesis 1.
The representations of $\mathcal{G}_{VAE}$ changes significantly on the first two tasks of simpler datasets (MNIST and FashionMNIST). These results indicate that it takes more time for the VAE model to learn stable features. In CIFAR-10, the amount of forgetting on VAE representations is almost equal to the amount of forgetting on AE representations which in both cases is almost negligible. This may suggest that the more complex the data is, the more general features the generative models learn.

Additionally, the dynamics of changes are less chaotic for the generative model, as the similarity of representations monotonically degrades with each task. This is in stark contrast to the discriminative representations which evolve chaotically. For instance, in the FashionMNIST dataset (middle column, top row), representations of the first task drastically change during the second and third task, then stabilize on the fourth task and unexpectedly become more similar during the last task. This may suggest that although the forgetting occurs in the generative case, it has a gradual form and is more predictable than in the discriminative part.

\subsection{Hypothesis 2}

To validate the second hypothesis, which states that \textit{autoencoders and variational autoencoders learn more transferable features than discriminative models}, we change the experimental protocol and learn the models $\mathcal{D}$, $\mathcal{G}_{AE}$ and $\mathcal{G}_{VAE}$ only on the first task in the learning sequence $T_{N}$. We adopt this approach as we are interested in measuring the transferability of learned features in the context of future tasks.

After the training for the first task is finished, we follow the procedure visualized in Fig~\ref{Fig:exp1_explanation} and collect representations for all data splits. Next, we train all classifiers on respective data splits. Then, we evaluate these classifiers on the validation datasets. Because the backbone of the neural network is trained only for the first task, we measure this way the transferability of the features learned during the first task to other tasks (results of this experiment are presented in the Table~\ref{Tab:after_first_task}). To gain further insights, we train a single classifier on all collected representations. This way, for each dataset, the task is a full 10-way classification. The results of this experiment tell us how general and useful are the features learned during the first task in the context of classifying the whole dataset. Results of this experiment are in the Table~\ref{Tab:acc_whole_dataset}.

\begin{table*}[t!]
  \centering
    \begin{tabular}{c||c|c|c}
    \toprule
    Model & MNIST & FashionMNIST & CIFAR-10\\
    \midrule
    {$\mathcal{D}$}& $76.8$ \textsubscript{$\pm 12.0$}  &  $74.3$ \textsubscript{$\pm 11.5$}   & $69.2$ \textsubscript{$\pm 6.4$}   \\
    %  & ($31.4$ \textsubscript{$\pm 3.1$}) & ($36.0$ \textsubscript{$\pm 3.0$}) & ($23.3$ \textsubscript{$\pm 2.3$}) \\
    \midrule
    {$\mathcal{G}_{AE}$} &\textbf{96.8 \textsubscript{$\pm$ 5.4}} &  92.1 \textsubscript{$\pm$ 7.2} & \textbf{80.5 \textsubscript{$\pm$ 2.0}} \\
    % & \textbf{(82.9 \textsubscript{$\pm$ 2.4})} & \textbf{(74.6 \textsubscript{$\pm$ 1.6})} & \textbf{ (44.4 \textsubscript{$\pm$ 1.0})} \\
    \midrule
    {$\mathcal{G}_{VAE}$} &88.9 \textsubscript{$\pm$ 6.7} & \textbf{94.2 \textsubscript{$\pm$ 5.5}}  & 78.9 \textsubscript{$\pm$ 3.0}  \\
    % & (58.6 \textsubscript{$\pm$ 5.2}) & (61.1 \textsubscript{$\pm$ 0.9}) & (37.4 \textsubscript{$\pm$ 0.2}) \\
    \bottomrule
    \end{tabular}
    \vspace{0.5cm}
    \caption{Average accuracy (in $\% \pm$ std) over all tasks after finishing the first task.}
    \label{Tab:after_first_task}
\end{table*}

\begin{table*}[t!]
  \centering
    \begin{tabular}{c||c|c|c}
    \toprule
    Model & MNIST & FashionMNIST & CIFAR-10\\
    \midrule
    % \multirow{2}{*}{$\mathcal{D}$}& $76.8$ \textsubscript{$\pm 12.0$}  &  $74.3$ \textsubscript{$\pm 11.5$}   & $69.2$ \textsubscript{$\pm 6.4$}   \\
    {$\mathcal{D}$} & $31.4$ \textsubscript{$\pm 3.1$} & $36.0$ \textsubscript{$\pm 3.0$} & $23.3$ \textsubscript{$\pm 2.3$} \\
    \midrule
    % \multirow{2}{*}{$\mathcal{G}_{AE}$} &\textbf{96.8 \textsubscript{$\pm$ 5.4}} &  92.1 \textsubscript{$\pm$ 7.2} & \textbf{80.5 \textsubscript{$\pm$ 2.0}} \\
    {$\mathcal{G}_{AE}$} & \textbf{82.9 \textsubscript{$\pm$ 2.4}} & \textbf{74.6 \textsubscript{$\pm$ 1.6}} & \textbf{ 44.4 \textsubscript{$\pm$ 1.0}} \\
    \midrule
    % \multirow{2}{*}{$\mathcal{G}_{VAE}$} &88.9 \textsubscript{$\pm$ 6.7} & \textbf{94.2 \textsubscript{$\pm$ 5.5}}  & 78.9 \textsubscript{$\pm$ 3.0}  \\
    {$\mathcal{G}_{VAE}$} & 58.6 \textsubscript{$\pm$ 5.2} & 61.1 \textsubscript{$\pm$ 0.9} & 37.4 \textsubscript{$\pm$ 0.2} \\
    \bottomrule
    \end{tabular}
    \vspace{0.5cm}
    \caption{Average accuracy (in $\% \pm$ std) for classification on the whole validation dataset (without splitting for tasks).}
    \label{Tab:acc_whole_dataset}
\end{table*}

% \begin{table*}[t!]
%   \centering
%   \begin{tabular}{c||c|c|c}
%     \toprule
%     Model & MNIST & FashionMNIST & CIFAR-10\\
%     \midrule
%     \multirow{2}{*}{$\mathcal{D}$}& $76.8$ \textsubscript{$\pm 12.0$}  &  $74.3$ \textsubscript{$\pm 11.5$}   & $69.2$ \textsubscript{$\pm 6.4$}   \\
%      & ($31.4$ \textsubscript{$\pm 3.1$}) & ($36.0$ \textsubscript{$\pm 3.0$}) & ($23.3$ \textsubscript{$\pm 2.3$}) \\
%     \midrule
%     \multirow{2}{*}{$\mathcal{G}_{AE}$} &\textbf{96.8 \textsubscript{$\pm$ 5.4}} &  92.1 \textsubscript{$\pm$ 7.2} & \textbf{80.5 \textsubscript{$\pm$ 2.0}} \\
%     & \textbf{(82.9 \textsubscript{$\pm$ 2.4})} & \textbf{(74.6 \textsubscript{$\pm$ 1.6})} & \textbf{ (44.4 \textsubscript{$\pm$ 1.0})} \\
%     \midrule
%     \multirow{2}{*}{$\mathcal{G}_{VAE}$} &88.9 \textsubscript{$\pm$ 6.7} & \textbf{94.2 \textsubscript{$\pm$ 5.5}}  & 78.9 \textsubscript{$\pm$ 3.0}  \\
%     & (58.6 \textsubscript{$\pm$ 5.2}) & (61.1 \textsubscript{$\pm$ 0.9}) & (37.4 \textsubscript{$\pm$ 0.2}) \\
%     \bottomrule
%   \end{tabular}
% \caption{Average accuracy (in $\% \pm$ std) over all tasks after finishing the first task. In brackets the accuracy for classification on the whole dataset (without splitting for tasks).}
% \label{Tab:after_first_task}
% \end{table*}

The superiority of the generative representations in terms of transferability is visible on all considered benchmarks. Although classifiers learned on $\mathcal{G}_{VAE}$ representations from the VAE model performs worse than classifiers trained on $\mathcal{G}_{AE}$ representations, both of these approaches obtain significantly better results than classifiers trained of model $\mathcal{D}$ representations. In the case of the model $\mathcal{G}_{AE}$ for MNIST and FashionMNIST datasets, the classifiers have a nearly perfect average accuracy of $96.8\%$ and $92.1\%$, respectively. The results of the classifiers trained on joint datasets (displayed in the brackets in Table~\ref{Tab:after_first_task}) are even more surprising. It turns out that the representations learned by the $\mathcal{G}_{AE}$ model on one task generalize well beyond that one task, which is proved by average accuracy of 82.9\%, 74.6\%, and 44.36\% in 10-way classification tasks on MNIST, FashionMNIST, and CIFAR-10 respectively. On the other hand, the poor performance of classifiers trained with the representations learned by the discriminative model confirms that representations learned this way do not generalize beyond the current task. This poor performance using representations of discriminative models is expected as the discriminative model aims to find the set of features that only separate the current data, resulting in useless features for upcoming tasks. We postulate that this poor transferability of features for upcoming tasks is the main reason for the catastrophic forgetting in discriminative models.

Guided by the above results, we create an additional experiment where we analyze how the training on the first task impacts the reconstruction quality on other tasks.
To that end, we carry out experiments only with the generative models $\mathcal{G}_{AE}$ and $\mathcal{G}_{VAE}$. During training on the first task, we evaluate the model on the reconstruction task with data samples from other tasks. In contrast to previous experiments, we do not perform any additional finetuning before evaluating the model. To evaluate the model on different tasks, we feed the data from this task and compute the corresponding reconstruction loss.

\begin{figure}[t!]
     \centering
     \includegraphics[width=\linewidth]{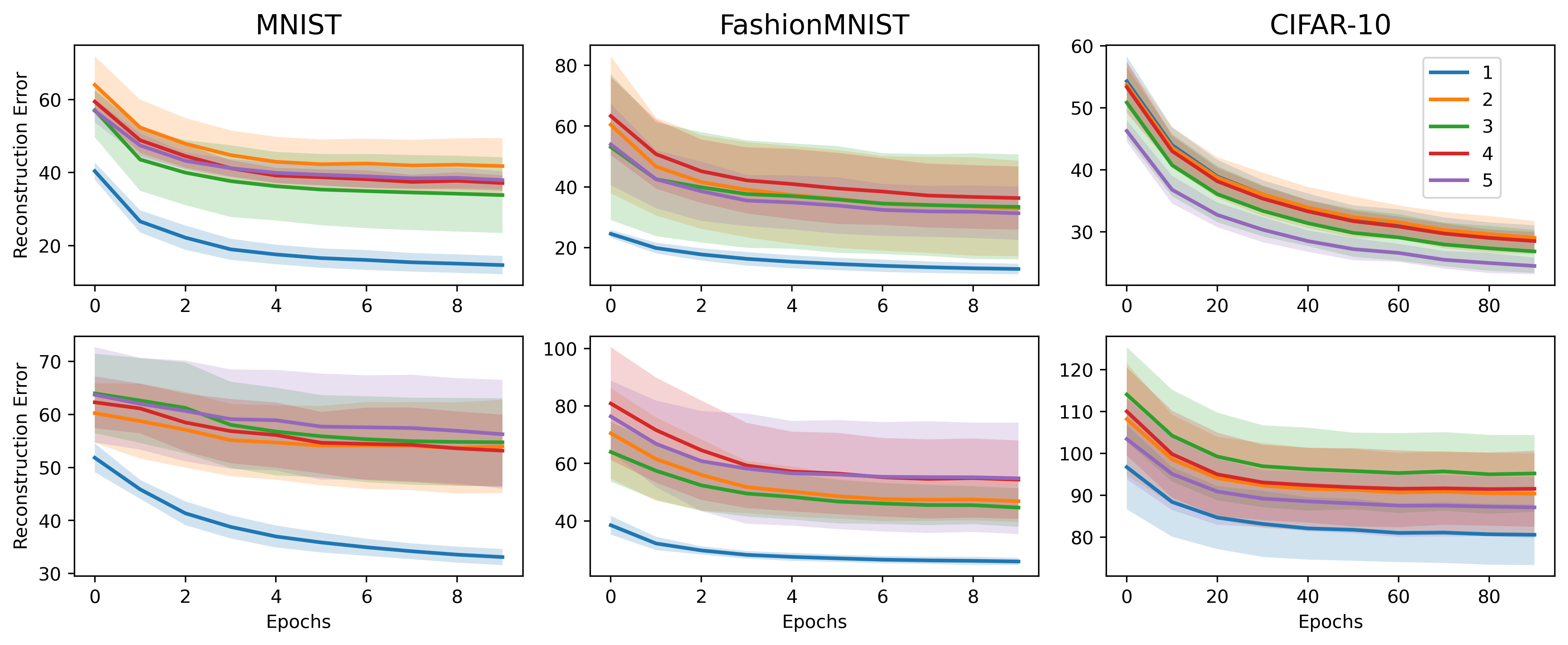}
      \caption{Reconstruction loss on different tasks for SplitMNIST (left), FashionMNIST (middle) and CIFAR-10 (right) with respect to training epochs on first task. Top row -- model $\mathcal{G}_{AE}$, bottom row -- $\mathcal{G}_{VAE}$. Results are averaged over 5 independent runs. The shaded area represents the standard deviation.}
      \label{Fig:reconstruction_loss}
\end{figure}

Fig.~\ref{Fig:reconstruction_loss} presents the results of this experiment. The top row presents the results for $\mathcal{G}_{AE}$ for MNIST, FashionMNIST, and CIFAR-10, respectively. The bottom row presents results for $\mathcal{G}_{VAE}$ with the same dataset ordering.
In each plot, the blue curve represents the loss on the first task -- the one that the model learns. All the plots have the same interesting tendency -- the reconstruction loss on all tasks decreases proportionally with the loss of the first task. This shows that the unvisited tasks directly benefit from the training on the first task. In other words, the representations learned during the first task are useful for the following tasks. 

In MNIST and FashionMNIST, there is a significant gap between the loss on the current task (blue curve) and other tasks. The gap is more significant in the case of $\mathcal{G}_{VAE}$, which is in line with the results from previous experiments suggesting that VAE based models need more complex datasets to develop general features. This is clearly visible in the rightmost column presenting results for CIFAR-10, where this gap does not exist. This may seem counter-intuitive as CIFAR-10 is the most challenging dataset in this group. However, the more complex the training data is, the more general the features learned by the generative model are. 

With these experiments, we provide an intuitive explanation to the phenomenon observed by Thai et al.~\cite{thai2021continual}, where authors suggest that the continual reconstruction task does not suffer from catastrophic forgetting. We argue that this is thanks to the generality of features learned by the generative model during the first task since the features learned by the autoencoder during the first task are also useful for the upcoming tasks. As can be observed in Fig~\ref{Fig:reconstruction_loss}) the model directly benefits from the learned features on previous tasks, and thus there is no interference between them, which results in the lack of catastrophic forgetting.

\section{Discussion}

In this work, we stated two hypotheses concerning different representations and their impact on catastrophic forgetting. We carried out experiments to empirically validate our hypotheses. 
From the performed experiments, it follows that the dynamics of the catastrophic forgetting for discriminative and generative representations are different. In particular, results from Fig.~\ref{Fig:cka} show that the forgetting of generative representations is gradual and monotonic. These properties suggest that it might be possible to model the effect of catastrophic forgetting in generative models precisely. However, to that end, one has to measure the effects of catastrophic forgetting exactly. Currently used proxy tasks for that purpose, such as classification tasks, hinder the analysis. Since it becomes more evident that catastrophic forgetting is not a homogeneous phenomenon and its character depends on the task or learning paradigm, we foresee the need for task agnostic measures of catastrophic forgetting. Such work would greatly benefit the community and give us unprecedented insights into the nature of the phenomenon.

From the practical point of view, this paper offers a new understanding of the dynamics of autoencoders and variational autoencoders learning in the continual scenario and their ability to generalize obtained knowledge to future tasks. These results can be directly adapted to the currently used generative rehearsal methods making them more robust or even immune to catastrophic forgetting. 

Last, this work is the first step towards a deeper understanding of the similarities and dissimilarities of generative and discriminative representations. While several papers discuss this relation, \textit{e.g.,}~\cite{wu2018tale} many questions remain unanswered, especially in the context of continual learning. How can one use the mechanisms of generative learning to obtain better representations for discrimination tasks in continual learning? Is the generic nature of generative representations the only explanation for the lack of catastrophic forgetting? Answering these and other questions, we left as future work.

\bibliographystyle{splncs04}
\bibliography{main}
\end{document}